%% file: main.tex
\documentclass{article}

\PassOptionsToPackage{numbers,compress}{natbib}
\usepackage[final]{neurips_2021}

\usepackage[utf8]{inputenc} 
\usepackage[T1]{fontenc}    
\usepackage{hyperref}       
\usepackage{url}            
\usepackage{booktabs}       
\usepackage{amsfonts}       
\usepackage{nicefrac}       
\usepackage{microtype}      

\input{math_commands.tex}

\usepackage{enumitem}
\usepackage{wrapfig}
\usepackage{multirow}
\usepackage{amsthm}
\usepackage{xcolor,colortbl}
\usepackage{subcaption}
\usepackage{graphicx}
\usepackage{amssymb,amsmath,amsthm}
\usepackage{caption}

\newcommand{\trans}{^{\top}}

\newtheorem{theorem}{Theorem}
\newtheorem{definition}{Definition}
\newtheorem{lemma}{Lemma}
\newtheorem{proposition}{Proposition}
\newtheorem{corollary}{Corollary}

\title{Labeling Trick: A Theory of Using Graph Neural Networks for Multi-Node Representation Learning}

\author{%
  Muhan Zhang$^{1,2,}$\thanks{Corresponding author: Muhan Zhang (\texttt{muhan@pku.edu.cn}). Work done as a research scientist at Facebook.}~~~~Pan Li$^{3,}$\thanks{Pan Li inspires the study of labeling tricks, proves Theorem~\ref{thm:link-aso-num}, and helps check the theoretical framework.}~~~~Yinglong Xia$^{4}$~~~~Kai Wang$^{4}$~~~~Long Jin$^{4}$\\
  ${}^1$Institute for Artificial Intelligence, Peking University\\ 
  ${}^2$Beijing Institute for General Artificial Intelligence\\ 
  ${}^3$Department of Computer Science, Purdue University\\ 
  ${}^4$Facebook AI\\
}

\begin{document}

\maketitle

\begin{abstract}

In this paper, we provide a theory of using graph neural networks (GNNs) for multi-node representation learning (where we are interested in learning a representation for a set of more than one node, such as link). We know that GNN is designed to learn single-node representations. When we want to learn a node set representation involving multiple nodes, a common practice in previous works is to directly aggregate the single-node representations obtained by a GNN into a joint node set representation. In this paper, we show a fundamental constraint of such an approach, namely the inability to capture the dependence between nodes in the node set, and argue that directly aggregating individual node representations does not lead to an effective joint representation for multiple nodes. Then, we notice that a few previous successful works for multi-node representation learning, including SEAL, Distance Encoding, and ID-GNN, all used node labeling. These methods first label nodes in the graph according to their relationships with the target node set before applying a GNN. Then, the node representations obtained in the labeled graph are aggregated into a node set representation. By investigating their inner mechanisms, we unify these node labeling techniques into a single and most general form---\textit{labeling trick}. We prove that with labeling trick a sufficiently expressive GNN learns the most expressive node set representations, thus in principle solves any joint learning tasks over node sets. Experiments on one important two-node representation learning task, link prediction, verified our theory. Our work explains the superior performance of previous node-labeling-based methods, and establishes a theoretical foundation of using GNNs for multi-node representation learning.

\end{abstract}

\section{Introduction}
Graph neural networks (GNNs)~\citep{scarselli2009graph,bruna2013spectral,duvenaud2015convolutional,li2015gated,kipf2016semi,defferrard2016convolutional,dai2016discriminative,velivckovic2017graph,zhang2018end,ying2018hierarchical} have achieved great successes in recent years. While GNNs have been well studied for single-node tasks (such as node classification) and whole-graph tasks (such as graph classification), using GNNs to predict a set of multiple nodes is less studied and less understood. Among such \textit{multi-node representation learning} problems, \textit{link prediction} (predicting the link existence/class/value between a set of two nodes) is perhaps the most important one due to its wide applications in practice, including friend recommendation in social networks~\citep{adamic2003friends}, movie recommendation in Netflix~\citep{bennett2007netflix}, protein interaction prediction~\citep{qi2006evaluation}, drug response prediction~\citep{stanfield2017drug}, knowledge graph completion~\citep{nickel2015review}, etc.
In this paper, we use link prediction as a medium to study GNN's multi-node representation learning ability. Note that although our examples and experiments are all around link prediction, our theory applies generally to all multi-node representation learning problems such as triplet~\citep{liu2021neural}, motif~\citep{besta2021motif} and subgraph~\citep{alsentzer2020subgraph} prediction tasks.



There are two main classes of GNN-based link prediction methods: Graph AutoEncoder (GAE)~\citep{kipf2016variational} and SEAL~\citep{zhang2018link,li2020distance}.
\textbf{GAE} (and its variational version VGAE~\citep{kipf2016variational}) first applies a GNN to the entire network to compute a representation for each node. The representations of the two end nodes of the link are then aggregated to predict the target link. GAE represents a common practice of using GNNs to learn multi-node representations. That is, first obtaining individual node representations through a GNN as usual, and then aggregating the representations of those nodes of interest as the multi-node representation. On the contrary, \textbf{SEAL} applies a GNN to an enclosing subgraph around each link, where nodes in the subgraph are \textit{labeled differently} according to their distances to the two end nodes before applying the GNN. Despite both using GNNs for link prediction, SEAL often shows much better practical performance than GAE. As we will see, the key lies in SEAL's \textbf{node labeling} step.



\begin{wrapfigure}[12]{L}{0.4\textwidth}
\centering
\vspace{-10pt}
\includegraphics[width=0.25\textwidth]{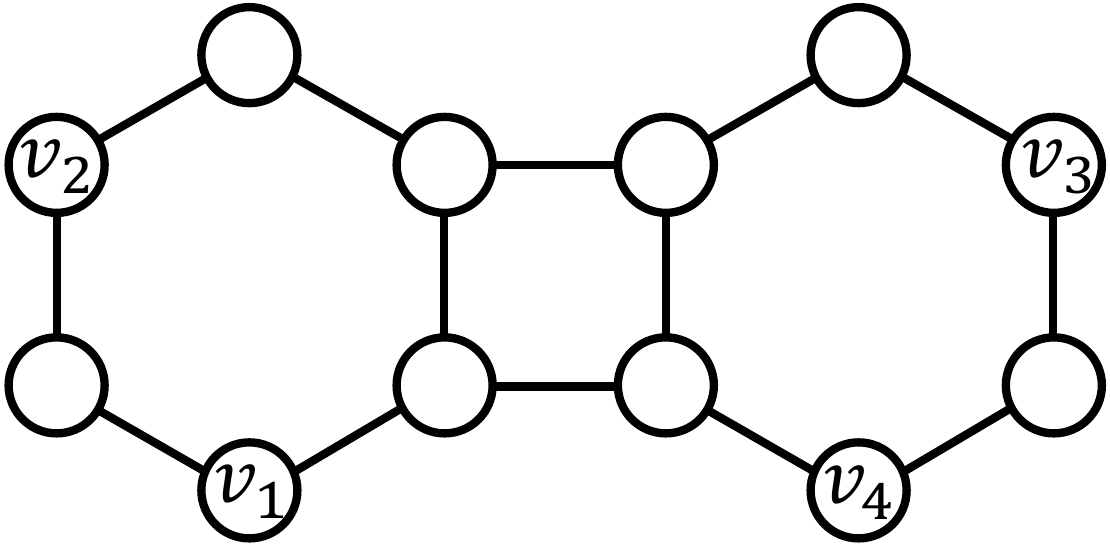}
\vspace{-5pt}
\caption{\small In this graph, nodes $v_2$ and $v_3$ are isomorphic; links $(v_1,v_2)$ and $(v_4,v_3)$ are isomorphic; link $(v_1,v_2)$ and link $(v_1,v_3)$ are \textbf{not} isomorphic. However, if we aggregate two node representations learned by a GNN as the link representation, we will give $(v_1,v_2)$ and $(v_1,v_3)$ the same prediction.}
\label{node_iso}
\end{wrapfigure}


We first give a simple example to show when GAE fails. In Figure~\ref{node_iso}, $v_2$ and $v_3$ have symmetric positions in the graph---from their respective views, they have exactly the \textbf{same $h$-hop neighborhood} for any $h$. Thus, without node features, GAE will learn the \textbf{same} representation for $v_2$ and $v_3$. However, when we want to predict which one of $v_2$ and $v_3$ is more likely to form a link with $v_1$, GAE will aggregate the representations of $v_1$ and $v_2$ as the link representation of $(v_1,v_2)$, and aggregate the representations of $v_1$ and $v_3$ to represent $(v_1,v_3)$, thus giving $(v_1,v_2)$ and $(v_1,v_3)$ the same representation and prediction. The failure to distinguish links $(v_1,v_2)$ and $(v_1,v_3)$ that have apparently different structural roles in the graph reflects one key limitation of GAE-type methods: by computing $v_1$ and $v_2$'s representations \textbf{independently} of each other, GAE cannot capture the \textbf{dependence} between two end nodes of a link. For example, $(v_1,v_2)$ has a much shorter path between them than that of $(v_1,v_3)$; and $(v_1,v_2)$ has both nodes in the same hexagon, while $(v_1,v_3)$ does not.


Take common neighbor (CN)~\citep{liben2007link}, one elementary heuristic feature for link prediction, as another example. CN counts the number of common neighbors between two nodes to measure their likelihood of forming a link, which is widely used in social network friend recommendation. CN is the foundation of many other successful heuristics such as Adamic-Adar~\citep{adamic2003friends} and Resource Allocation~\citep{zhou2009predicting}, which are also based on neighborhood overlap. However, GAE \textbf{cannot} capture such neighborhood-overlap-based features. This can be seen from Figure~\ref{node_iso} too. There is 1 common neighbor between $(v_1,v_2)$ and 0 between $(v_1,v_3)$, but GAE always gives $(v_1,v_2)$ and $(v_1,v_3)$ the same representation. The failure to learn common neighbor demonstrates GAE's severe limitation for link prediction. The root cause still lies in that GAE computes node representations independently of each other---when computing the representation of one end node, it is \textbf{not aware} of the other end node.



One way to alleviate the above failure is to use one-hot encoding of node indices or random features as input node features~\citep{loukas2019graph,sato2020random}. With such node-discriminating features, $v_2$ and $v_3$ will have different node representations, thus $(v_1,v_2)$ and $(v_1,v_3)$ may also have different link representations after aggregation, enabling GAE to discriminate $(v_1,v_2)$ and $(v_1,v_3)$. However, using node-discriminating features loses GNN's \textbf{inductive learning} ability to map nodes and links with identical neighborhoods (such as nodes $v_2$ and $v_3$, and links $(v_1,v_2)$ and $(v_4,v_3)$) to the same representation, which results in a great loss of generalization ability. The resulting model is no longer permutation invariant/equivariant, violating the fundamental design principle of GNNs. Is there a way to improve GNNs' link discriminating power (so that links like $(v_1,v_2)$ and $(v_1,v_3)$ can be distinguished), while maintaining their inductive learning ability (so that links $(v_1,v_2)$ and $(v_4,v_3)$ have the same representation)?



In this paper, we analyze the above problem from a \textit{structural representation learning} point of view. \citet{Srinivasan2020On} prove that the multi-node prediction problem on graphs ultimately only requires finding a \textit{most expressive structural representation} of node sets, which gives two node sets the same representation if and only if they are \textit{isomorphic} (a.k.a. symmetric, on the same orbit) in the graph. For example, link $(v_1, v_2)$ and link $(v_4, v_3)$ in Figure~\ref{node_iso} are isomorphic. A most expressive structural representation for links should give any two isomorphic links the same representation while discriminating all non-isomorphic links (such as $(v_1, v_2)$ and $(v_1, v_3)$). According to our discussion above, GAE-type methods that directly aggregate node representations cannot learn a most expressive structural representation. 
Then, how to learn a most expressive structural representation of node sets?

To answer this question, we revisit the other GNN-based link prediction framework, SEAL, and analyze how node labeling helps a GNN learn better node set representations. We find out that two properties of a node labeling are crucial for its effectiveness: 1) \textit{target-nodes-distinguishing} and 2) \textit{permutation equivariance}. With these two properties, we define \textit{labeling trick} (Section~\ref{sec:labelingtrick}), which unifies previous node labeling methods into a single and most general form. 
Theoretically, we prove that with labeling trick a sufficiently expressive GNN can learn most expressive structural representations of node sets (Theorem~\ref{labeltrickthm}), which reassures GNN's node set prediction ability. It also closes the gap between GNN's node representation learning nature and node set tasks' multi-node representation learning requirement. 
We further extend our theory to \textit{local isomorphism} (Section~\ref{localiso}).
And finally, experiments on four OGB link existence prediction datasets~\citep{hu2020open} verified our theory.

Note that the labeling trick theory allows the presence of node/edge features/types, thus is not restricted to non-attributed and homogeneous graphs. Previous works on heterogeneous graphs, such as knowledge graphs~\citep{teru2020inductive} and recommender systems~\citep{Zhang2020Inductive} have already seen successful applications of labeling trick. Labeling trick is also not restricted to two-node link representation learning tasks, but generally applies to any multi-node representation learning tasks.

\section{Preliminaries}
In this section, we introduce some important concepts that will be used in the analysis of the paper, including \textit{permutation}, \textit{set isomorphism} and \textit{most expressive structural representation}.


We consider a graph $\gG=(V,E,\tA)$, where $V=\{1,2,\ldots,n\}$ is the set of $n$ vertices, $E \subseteq V\times V$ is the set of edges, and $\tA \in \R^{n\times n \times k}$ is a 3-dimensional tensor containing node and edge features. The diagonal components $\tA_{i,i,:}$ denote features of node $i$, and the off-diagonal components $\tA_{i,j,:}$ denote features of edge $(i,j)$. For heterogeneous graphs, the node/edge types can also be expressed in $\tA$ using integers or one-hot encoding vectors. We further use $\mA \in \{0,1\}^{n\times n}$ to denote the adjacency matrix of $\gG$ with $\mA_{i,j}=1$ iff $(i,j)\in E$. 
We let $\mA$ be the first slice of $\tA$, i.e., $\mA = \tA_{:,:,1}$. Since $\tA$ contains the complete information of a graph, we sometimes directly use $\tA$ to denote the graph.

\begin{definition}\label{def:permutation} A \textbf{permutation} $\pi$ is a bijective mapping from $\{1,2,\ldots,n\}$ to $\{1,2,\ldots,n\}$. Depending on the context, $\pi(i)$ can mean assigning a new index to node $i\in V$, or mapping node $i$ to node $\pi(i)$ of another graph. All $n!$ possible $\pi$'s constitute the permutation group $\Pi_n$. 
For joint prediction tasks over a set of nodes, we use $S$ to denote the \textbf{target node set}. For example, $S=\{i,j\}$ if we want to predict the link between $i,j$. We define $\pi(S) = \{\pi(i) | i\in S\}$. We further define the permutation of $\tA$ as $\pi(\tA)$, where $\pi(\tA)_{\pi(i),\pi(j),:} = \tA_{i,j,:}$. 
\end{definition}

Next, we define \textit{set isomorphism}, which generalizes graph isomorphism to arbitrary node sets.
\begin{definition}\label{set_iso}
\textbf{(Set isomorphism)} Given two $n$-node graphs $\gG=(V,E,\tA)$, $\gG'=(V',E',\tA')$, and two node sets $S\subseteq V$, $S'\subseteq V'$, we say $(S,\tA)$ and $(S',\tA')$ are isomorphic (denoted by $(S,\tA) \simeq (S',\tA')$) if ~$\exists \pi \in \Pi_n$ such that $S = \pi(S')$ and $\tA = \pi(\tA')$. 
\end{definition}
When $(V,\tA) \simeq (V',\tA')$, we say two graphs $\gG$ and $\gG'$ are \textit{isomorphic} (abbreviated as $\tA \simeq \tA'$ because $V = \pi(V')$ for any $\pi$). 
Note that set isomorphism is \textbf{more strict} than graph isomorphism, because it not only requires graph isomorphism, but also requires the permutation maps a specific node set $S$ to another node set $S'$. 
In practice, when $S\neq V$, we are often more concerned with the case of $\tA=\tA'$, where isomorphic node sets are defined \textbf{in the same graph} (automorphism). For example, when $S = \{i\}, S' = \{j\}$ and $(i,\tA) \simeq (j,\tA)$, we say nodes $i$ and $j$ are isomorphic in graph $\tA$ (or they have symmetric positions/same structural role in graph $\tA$). An example is $v_2$ and $v_3$ in Figure~\ref{node_iso}.

We say a function $f$ defined over the space of $(S,\tA)$ is \textit{permutation invariant} (or \textit{invariant} for abbreviation) if $\forall \pi \in \Pi_n$, $f(S,\tA) = f(\pi(S),\pi(\tA))$. Similarly, $f$ is \textit{permutation equivariant} if $\forall \pi \in \Pi_n$, $\pi(f(S,\tA)) = f(\pi(S),\pi(\tA))$. Permutation invariance/equivariance ensures representations learned by a GNN is invariant to node indexing, which is a fundamental design principle of GNNs.




Now we define most expressive structural representation of a node set, following \citep{Srinivasan2020On,li2020distance}. Basically, it assigns a unique representation to each equivalence class of isomorphic node sets.

\begin{definition}\label{structuralrepresentation}
Given an invariant function $\Gamma(\cdot)$, $\Gamma(S,\tA)$ is a \textbf{most expressive structural representation} for $(S,\tA)$ if ~$\forall S,\tA, S',\tA', ~\Gamma(S,\tA) = \Gamma(S',\tA') \Leftrightarrow (S,\tA) \simeq (S',\tA')$.
\end{definition}

For simplicity, we will briefly use \textit{structural representation} to denote most expressive structural representation in the rest of the paper. We will omit $\tA$ if it is clear from context. We call $\Gamma(i,\tA)$ a \textit{structural node representation} for $i$, and call $\Gamma(\{i,j\},\tA)$ a \textit{structural link representation} for $(i,j)$. 


Definition~\ref{structuralrepresentation} requires that the structural representations of two node sets are the same if and only if the two node sets are isomorphic. That is, isomorphic node sets always have the \textbf{same} structural representation, while non-isomorphic node sets always have \textbf{different} structural representations.
This is in contrast to \textit{positional node embeddings} such as DeepWalk~\citep{perozzi2014deepwalk} and matrix factorization~\citep{mnih2008probabilistic}, where two isomorphic nodes can have different node embeddings~\citep{ribeiro2017struc2vec}. GAE using node-discriminating features also learns positional node embeddings.



\textbf{Why do we study structural representations?} Formally speaking, \citet{Srinivasan2020On} prove that any joint prediction task over node sets only requires a structural representation of node sets. They show that positional node embeddings carry no more information beyond that of structural representations. Intuitively speaking, it is because two isomorphic nodes in a network are perfectly symmetric and interchangeable with each other, and should be indistinguishable from any perspective. Learning a structural node representation guarantees that isomorphic nodes are always classified into the same class. Similarly, learning a structural link representation guarantees isomorphic links, such as $(v_1, v_2)$ and $(v_4, v_3)$ in Figure~\ref{node_iso}, are always predicted the same, while non-isomorphic links, such as $(v_1, v_2)$ and $(v_1, v_3)$, are always distinguishable, which is not guaranteed by positional node embeddings. Structural representation characterizes the maximum representation power a model can reach on graphs. We use it to study GNNs' multi-node representation learning ability. 

\section{The limitation of directly aggregating node representations}
In this section, using GAE for link prediction as an example, we show the key limitation of directly aggregating node representations as a node set representation.


\setlength{\abovedisplayskip}{5pt}
\setlength{\belowdisplayskip}{5pt}

\subsection{GAE for link prediction}
Given a graph $\tA$, GAE methods~\citep{kipf2016variational} first use a GNN to compute a node representation $\vz_i$ for each node $i$, and then use an aggregation function $f(\{\vz_i,\vz_j\})$ to predict link $(i,j)$:
\begin{align}
    \hat{\mA}_{i,j} = f(\{\vz_i,\vz_j\}), ~\text{where}~\vz_i \!=\! \text{GNN}(i,\tA), \vz_j \!=\! \text{GNN}(j,\tA).\nonumber
\end{align}
Here $\hat{\mA}_{i,j}$ is the predicted score for link $(i,j)$. The model is trained to maximize the likelihood of reconstructing the true adjacency matrix. The original GAE uses a two-layer GCN~\citep{kipf2016semi} as the GNN, and let $f(\{\vz_i,\vz_j\}):=\sigma(\vz_i\trans \vz_j)$. In principle, we can replace GCN with any GNN, and replace $\sigma(\vz_i\trans \vz_j)$ with an MLP over any aggregation function over $\{\vz_i,\vz_j\}$. Besides inner product, other aggregation choices include mean, sum, bilinear product, concatenation, and Hadamard product. In the following, we will use GAE to denote a general class of GNN-based link prediction methods. 

GAE uses a GNN to learn node representations and then aggregates pairwise node representations as link representations. Two natural questions to ask are: 1) Is the node representation learned by the GNN a \textit{structural node representation}? 2) Is the link representation aggregated from two node representations a \textit{structural link representation}? We answer them respectively in the following.

\subsection{GNN and structural node representation}
Practical GNNs~\citep{gilmer2017neural} usually simulate the 1-dimensional Weisfeiler-Lehman (1-WL) test~\citep{weisfeiler1968reduction} to iteratively update each node's representation by aggregating its neighbors' representations. We use \textit{1-WL-GNN} to denote a GNN with 1-WL discriminating power, such as GIN~\citep{xu2018powerful}.

A 1-WL-GNN ensures that isomorphic nodes always have the same representation. But the opposite direction is not guaranteed. For example, a 1-WL-GNN gives the same representation to all nodes in an $r$-regular graph. Despite this, 1-WL is known to discriminate almost all non-isomorphic nodes~\citep{babai1979canonical}. This indicates that a 1-WL-GNN can always give the same representation to isomorphic nodes, and can give different representations to \textbf{almost all} non-isomorphic nodes.


To study GNN's maximum expressive power for multi-node representation learning, we also define a \textit{node-most-expressive GNN}, which gives different representations to \textbf{all} non-isomorphic nodes.

\begin{definition}
A \text{\normalfont GNN} is \textbf{node-most-expressive} if ~$\forall i,\tA$,$j,\tA'$, $~\text{\normalfont GNN}(i,\tA) = \text{\normalfont GNN}(j,\tA') \Leftrightarrow (i,\tA) \simeq (j,\tA')$.
\end{definition}
That is, node-most-expressive GNN learns \textit{structural node representations}\footnote{Although a polynomial-time implementation is not known for node-most-expressive GNNs, many practical softwares can discriminate all non-isomorphic nodes quite efficiently~\citep{mckay2014practical}, which provides a promising direction.}. We define such a GNN because we want to answer: whether GAE, even equipped with a node-most-expressive GNN (so that GNN's node representation power is not a bottleneck), can learn structural link representations.

\subsection{GAE \textbf{cannot} learn structural link representations}\label{gae_limitation}

Suppose GAE is equipped with a node-most-expressive GNN which outputs structural node representations. 
Then the question becomes: does the aggregation of structural node representations of $i$ and $j$ result in a structural \textit{link} representation of $(i,j)$? The answer is no, as shown in previous works~\citep{Srinivasan2020On,Zhang2020Inductive}. 
We have also illustrated it in the introduction:
In Figure~\ref{node_iso}, we have two isomorphic nodes $v_2$ and $v_3$, thus $v_2$ and $v_3$ will have the same structural node representation. By aggregating structural node representations, GAE will give $(v_1,v_2)$ and $(v_1,v_3)$ the same link representation. However, $(v_1,v_2)$ and $(v_1,v_3)$ are not isomorphic in the graph. This indicates:

\begin{proposition}
GAE \textbf{cannot} learn structural link representations no matter how expressive node representations a GNN can learn.
\end{proposition}

Similarly, we can give examples like Figure~\ref{node_iso} for multi-node representation learning problems involving more than two nodes to show that directly aggregating node representations from a GNN does not lead to a structural representation for node sets. 
The root cause of this problem is that GNN computes node representations independently, without being aware of the other nodes in the target node set $S$. Thus, even GNN learns the most expressive single-node representations, there is never a guarantee that their aggregation is a structural representation of a node set. In other words, the multi-node representation learning problem is \textbf{not breakable} into multiple \textbf{independent} single-node representation learning problems. If we have to break it, the multiple single-node representation learning problems should be \textbf{dependent} on each other.





\section{Labeling trick for multi-node representation learning}

In this section, we first define the general form of \textit{labeling trick}, and use a specific implementation, zero-one labeling trick, to intuitively explain why labeling trick helps GNNs learn better link representations. Next, we present our main theorem showing that labeling trick enables a node-most-expressive GNN to learn structural representations of node sets, which formally characterizes GNN's maximum multi-node representation learning ability. Then, we review SEAL and show it exactly uses one labeling trick. Finally, we discuss other labeling trick implementations in previous works. 





\subsection{Labeling trick}\label{sec:labelingtrick}

\begin{definition}\label{labelingtrick}
\textbf{(Labeling trick)} Given $(S,\tA)$, we stack a labeling tensor $\tL^{(S)} \in \R^{n\times n\times d}$ in the third dimension of $\tA$ to get a new $\tA^{(S)} \in \R^{n\times n\times (k+d)}$, where $\tL$ satisfies: $\forall S,\tA, S',\tA',\pi \in \Pi_n$, \\
\text{\normalfont 1.} (\textit{target-nodes-distinguishing}) ~$\tL^{(S)} = \pi(\tL^{(S')}) \Rightarrow S = \pi(S')$, and\\
\text{\normalfont 2.} (\textit{permutation equivariance}) ~~$S = \pi(S'), \tA = \pi(\tA') \Rightarrow \tL^{(S)} = \pi(\tL^{(S')})$. 
\end{definition}


To explain a bit, labeling trick assigns a label vector to each node/edge in graph $\tA$, which constitutes the labeling tensor $\tL^{(S)}$. By concatenating $\tA$ and $\tL^{(S)}$, we get the new labeled graph $\tA^{(S)}$. By definition we can assign labels to both nodes and edges. However, in this paper, we \textbf{only consider node labels} for simplicity, i.e., we let the off-diagonal components $\tL^{(S)}_{i,j,:}$ be all zero.

The labeling tensor $\tL^{(S)}$ should satisfy two properties in Definition~\ref{labelingtrick}. Property 1 requires that if a permutation $\pi$ preserving node labels (i.e., $\tL^{(S)} = \pi(\tL^{(S')})$) exists between nodes of $\tA$ and $\tA'$, then the nodes in $S'$ must be mapped to nodes in $S$ by $\pi$ (i.e., $S = \pi(S')$). A sufficient condition for property 1 is to make the target nodes $S$ have \textit{distinct labels} from those of the rest nodes, so that $S$ is \textit{distinguishable} from others.
Property 2 requires that when $(S,\tA)$ and $(S',\tA')$ are isomorphic under $\pi$ (i.e., $S = \pi(S'), \tA = \pi(\tA')$), the corresponding nodes $i\in S, j\in S', i=\pi(j)$ must always have the same label (i.e., $\tL^{(S)} = \pi(\tL^{(S')})$). A sufficient condition for property 2 is to make the labeling function \textit{permutation equivariant}, i.e., when the target $(S,\tA)$ changes to $(\pi(S),\pi(\tA))$, the labeling tensor $\tL^{(S)}$ should equivariantly change to $\pi(\tL^{(S)})$.






Now we introduce a simplest labeling trick satisfying the two properties in Definition~\ref{labelingtrick}, and use it to illustrate how labeling trick helps GNNs learn better node set representations.

\begin{definition}\label{zolabeling}
\textbf{(Zero-one labeling trick)} Given a graph $\tA$ and a set of nodes $S$ to predict, we give it a diagonal labeling matrix $\tL^{(S)} \in \R^{n\times n\times 1}$ such that $\tL^{(S)}_{i,i,1} = 1$ if $i\in S$ and $\tL^{(S)}_{i,i,1} = 0$ otherwise.
\end{definition}


In other words, the zero-one labeling trick assigns label 1 to nodes in $S$, and label 0 to all nodes not in $S$. It is a valid labeling trick because firstly, nodes in $S$ get \textit{distinct labels}, and secondly, the labeling function is \textit{permutation equivariant} by always giving nodes in the target node set a label 1.
These node labels serve as additional node features fed to a GNN together with the original node features.





\begin{figure}[tp]
\centering
\includegraphics[width=0.58\textwidth]{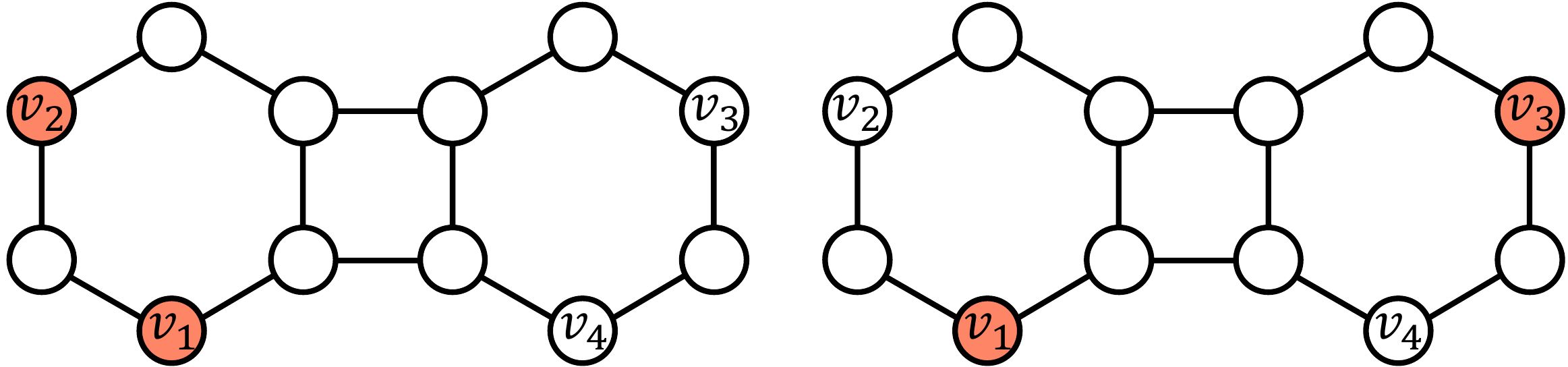}
\caption{\small When we predict $(v_1,v_2)$, we will label these two nodes differently from the rest, so that a GNN is aware of the target link when learning $v_1$ and $v_2$'s representations. Similarly, when predicting $(v_1,v_3)$, nodes $v_1$ and $v_3$ will be labeled differently. This way, the representation of $v_2$ in the left graph will be different from that of $v_3$ in the right graph, enabling GNNs to distinguish the non-isomorphic links $(v_1,v_2)$ and $(v_1,v_3)$.
}
\label{link_iso}
\end{figure}

Let's return to the example in Figure~\ref{node_iso} to see how the zero-one labeling trick helps GNNs learn better link representations. This time, when we want to predict link $(v_1,v_2)$, we will label $v_1,v_2$ differently from the rest nodes, as shown by the different color in Figure~\ref{link_iso}~left. With nodes $v_1$ and $v_2$ labeled, when the GNN is computing $v_2$'s representation, it is also ``aware'' of the source node $v_1$, instead of the previous agnostic way that treats $v_1$ the same as other nodes. Similarly, when we want to predict link $(v_1,v_3)$, we will again label $v_1,v_3$ differently from other nodes as shown in Figure~\ref{link_iso}~right. This way, $v_2$ and $v_3$'s node representations are no longer the same in the two differently labeled graphs (due to the presence of the labeled $v_1$), and we are able to predict $(v_1,v_2)$ and $(v_1,v_3)$ differently. The key difference from GAE is that the node representations are no longer computed independently, but are \textit{conditioned} on each other in order to capture the dependence between nodes.


At the same time, isomorphic links, such as $(v_1,v_2)$ and $(v_4,v_3)$, will still have the same representation, since the zero-one labeled graph for $(v_1,v_2)$ is still symmetric to the zero-one labeled graph for $(v_4,v_3)$. This brings an exclusive advantage over GAE using node-discriminating features. 

With $v_1$ and $v_2$ labeled, a GNN can also learn their common neighbor easily: in the first iteration, only $(v_1,v_2)$'s common neighbors will receive the distinct message from both $v_1$ and $v_2$; then in the next iteration, all common neighbors will pass their distinct messages back to both $v_1$ and $v_2$, which effectively encode the number of common neighbors into $v_1$ and $v_2$'s updated representations.


Now we introduce our main theorem showing that with a valid labeling trick, a node-most-expressive GNN can \textit{learn structural representations of node sets}. 

\begin{theorem}\label{labeltrickthm}
Given a node-most-expressive \text{\normalfont GNN} and an injective set aggregation function \text{\normalfont AGG}, for any $S,\tA, S',\tA'$, $\text{\normalfont GNN}(S,\tA^{(S)}) = \text{\normalfont GNN}(S',\tA'^{(S')}) \Leftrightarrow (S,\tA) \!\simeq\! (S',\tA') \nonumber$, where $\text{\normalfont GNN}(S,\tA^{(S)}) := \text{\normalfont AGG}(\{\text{\normalfont GNN}(i,\tA^{(S)}) | i\in S\})$.
\end{theorem}

We include all proofs in the appendix. 
Theorem~\ref{labeltrickthm} implies that $\text{\normalfont AGG}(\{\text{\normalfont GNN}(i,\tA^{(S)}) | i\in S\})$ is a structural representation for $(S,\tA)$. Remember that directly aggregating the structural node representations learned from the original graph $\tA$ does not lead to structural representations of node sets (Section~\ref{gae_limitation}). Theorem~\ref{labeltrickthm} shows that aggregating the structural node representations learned from the \textbf{labeled} graph $\tA^{(S)}$, somewhat surprisingly, results in a structural representation for $(S,\tA)$.

The significance of Theorem~\ref{labeltrickthm} is that it closes the gap between GNN's single-node representation nature and node set prediction problems' multi-node representation requirement. It demonstrates that GNNs are able to learn most expressive structural representations of node sets, thus are suitable for joint prediction tasks over node sets too. This answers the open question raised in \citep{Srinivasan2020On} questioning GNNs' link prediction ability: \textit{are structural node representations in general--and GNNs in particular--fundamentally incapable of performing link (dyadic) and multi-ary (polyadic) prediction tasks?} With Theorem~\ref{labeltrickthm}, we argue the answer is no. Although GNNs alone have severe limitations for learning joint representations of multiple nodes, GNNs + labeling trick can learn structural representations of node sets too by aggregating structural node representations obtained in the labeled graph.

Theorem~\ref{labeltrickthm} assumes a node-most-expressive GNN. To augment Theorem~\ref{labeltrickthm}, we give the following theorem, which demonstrates labeling trick's power for 1-WL-GNNs.

\begin{theorem} \label{thm:link-aso-num}
In any non-attributed graph with $n$ nodes, if the degree of each node in the graph is between $1$ and $\mathcal{O}(\log^{\frac{1-\epsilon}{2h}} n)$ for any constant $\epsilon>0$, then there exists $\omega(n^{2\epsilon})$ many pairs of non-isomorphic links $(u,w), (v,w)$ such that an $h$-layer \textnormal{1-WL-GNN} gives $u,v$ the same representation, while with labeling trick the \textnormal{1-WL-GNN} gives $u,v$ different representations.
\end{theorem}

Theorem~\ref{thm:link-aso-num} shows that in any non-attributed graph there exists a large number ($\omega(n^{2\epsilon})$) of link pairs (like the examples $(v_1,v_2)$ and $(v_1,v_3)$ in Figure~\ref{node_iso}) which are not distinguishable by 1-WL-GNNs alone but distinguishable by 1-WL-GNNs + labeling trick.






\subsection{SEAL uses a labeling trick}

SEAL~\citep{zhang2018link} is a state-of-the-art link prediction method based on GNNs. It first extracts an \textit{enclosing subgraph} ($h$-hop subgraph) around each target link to predict.

\begin{definition}
\textbf{(Enclosing subgraph)} Given $(S,\tA)$, the $h$-hop enclosing subgraph $\tA_{(S,h)}$ of $S$ is the subgraph induced from $\tA$ by $\cup_{j\in S} \{i ~|~ d(i,j) \leq h\}$, where $d(i,j)$ is the shortest path distance between nodes $i$ and $j$. 
\end{definition}

Then, SEAL applies Double Radius Node Labeling (DRNL) to give an integer label to each node in the enclosing subgraph. DRNL assigns different labels to nodes with \textbf{different distances} to the two end nodes of the link. It works as follows: The two end nodes are always labeled 1. Nodes farther away from the two end nodes get larger labels (starting from 2). For example, nodes with distances $\{1,1\}$ to the two end nodes will get label 2, and nodes with distances $\{1,2\}$ to the two end nodes will get label 3. So on and so forth. Finally the labeled enclosing subgraph is fed to a GNN to learn the link representation and output the probability of link existence. 

\begin{theorem}\label{drnlthm}
DRNL is a labeling trick.
\end{theorem}

Theorem~\ref{drnlthm} is easily proved by noticing: across different subgraphs, 1) nodes with label 1 are always those in the target node set $S$, and 2) nodes with the same distances to $S$ always have the same label, while distances are permutation equivariant. Thus, SEAL exactly uses a specific labeling trick to enhance its power, which explains its often superior performance than GAE~\citep{zhang2018link}. 

SEAL only uses a subgraph $\tA_{(S,h)}$ within $h$ hops from the target link instead of using the whole graph. This is not a constraint but rather a practical consideration (just like GAE typically uses less than 3 message passing layers in practice), and its benefits will be discussed in detail in Section~\ref{localiso}. When $h\to \infty$, the subgraph becomes the entire graph, and SEAL is able to \textit{learn structural link representations} from the labeled (entire) graph. 
\begin{proposition}
When $h\to\infty$, SEAL can learn structural link representations with a node-most-expressive GNN.
\end{proposition}


\subsection{Discussion}\label{sec:discussion}

\noindent\textbf{DE and DRNL}~~In~\citep{li2020distance}, SEAL's distance-based node labeling scheme is generalized to \textit{Distance Encoding} (DE) that can be applied to $|S| > 2$ problems. Basically, DRNL is equivalent to DE-2 using shortest path distance. Instead of encoding two distances into one integer label, DE injectively aggregates the embeddings of two distances into a label vector. \textbf{DE is also a valid labeling trick}, as it can also distinguish $S$ and is permutation equivariant. 
However, there are some subtle differences between DE and DRNL's implementations, which are discussed in Appendix~\ref{drnlde}.





\noindent\textbf{ID-GNN}~~\citet{you2021identity} propose \textit{Identity-aware GNN} (ID-GNN), which assigns a unique color to the ``identity'' nodes and performs message passing for them with a different set of parameters. ID-GNN's coloring scheme is similar to the zero-one labeling trick that distinguishes nodes in the target set with 0/1 labels. However, when used for link prediction, ID-GNN only colors the source node, while the zero-one labeling trick labels both the source and destination nodes. 
Thus, ID-GNN can be seen as using a partial labeling trick. The idea of conditioning on only the source node is also used in NBFNet~\citep{zhu2021neural}. We leave the exploration of partial labeling trick's power for future work.




\noindent\textbf{Labeling trick for heterogeneous graphs}~~Since our graph definition $\tA$ allows the presence of node/edge types, our theory applies to heterogeneous graphs, too. In fact, previous works have already successfully used labeling trick for heterogeneous graphs. IGMC~\citep{Zhang2020Inductive} uses labeling trick to predict ratings between users and items (recommender systems), where a user node $k$-hop away from the target link receives a label $2k$, and an item node $k$-hop away from the target link receives a label $2k+1$. It is a valid labeling trick since the target user and item always receive distinct labels 0 and 1. On the other hand, GRAIL~\citep{teru2020inductive} applies the DRNL labeling trick to knowledge graph completion.

\noindent\textbf{Directed case.}~~Despite that we do not restrict our graphs to be undirected, our node set definition (Definition~\ref{set_iso}) does not consider the order of nodes in the set (i.e., direction of link when $|S|=2$). The ordered case assumes $S=(1,2,3)$ is different from $S'=(3,2,1)$. One way to solve this is to define labeling trick respecting the order of $S$. In fact, if we define $\pi(S)=\big(\pi(S[i]) ~|~ i=1,2,\ldots,|S| \big)$ (where $S[i]$ denotes the $i^{\text{th}}$ element in the ordered set $S$) in Definition~\ref{def:permutation}, and modify our definition of labeling trick using this new definition of permutation, then Theorem~\ref{labeltrickthm} still holds.


\noindent\textbf{Complexity.}~~Despite the power, labeling trick may introduce extra computational complexity. The reason is that for every node set $S$ to predict, we need to relabel the graph $\tA$ according to $S$ and compute a new set of node representations within the labeled graph. In contrast, GAE-type methods compute node representations only in the original graph. For small graphs, GAE-type methods can compute all node representations first and then predict multiple node sets at the same time, which saves a significant amount of time. However, for large graphs that cannot fit into the GPU memory, mini-batch training (which extracts a neighborhood subgraph for every node set to predict) has to be used for both GAE-type methods and labeling trick, resulting in similar computation cost.

\section{Local isomorphism: a more practical view of isomorphism}\label{localiso}


The concept of most expressive structural representation is based on assigning node sets the same representation if and only if they are \textit{isomorphic} to each other in the graph. However, exact isomorphism is not very common. For example, \citet{babai1979canonical} prove that at least $(n-\log n)$ nodes in almost all $n$-node graphs are \textit{non-isomorphic} to each other. In practice, 1-WL-GNN also takes up to $\mathcal{O}(n)$ message passing layers to reach its maximum power for discriminating non-isomorphic nodes, making it very hard to really target on finding exactly isomorphic nodes/links.

\begin{lemma}\label{wlconverge}
Given a graph with $n$ nodes, a 1-WL-GNN takes up to $\mathcal{O}(n)$ message passing layers to discriminate all the nodes that 1-WL can discriminate. \end{lemma}

In this regard, we propose a more practical concept, called \textit{local isomorphism}. 

\begin{definition}\label{localhiso}
\textbf{(Local $h$-isomorphism)} $\forall S,\tA, S',\tA'$, we say $(S,\tA)$ and $(S',\tA')$ are locally $h$-isomorphic to each other if $(S,\tA_{(S,h)}) \simeq (S',\tA'_{(S',h)})$.
\end{definition}

Local $h$-isomorphism only requires the $h$-hop enclosing subgraphs around $S$ and $S'$ are isomorphic, instead of the entire graphs. We argue that this is a more useful definition than isomorphism, because: 1) Exact isomorphism is rare in real-world graphs. 2) Algorithms targeting on exact isomorphism are more likely to overfit. Only assigning the same representations to exactly isomorphic nodes/links may fail to identify a large amount of nodes/links that are not isomorphic but have very similar neighborhoods. Instead, nodes/links \textit{locally isomorphic} to each other may better indicate that they should have the same representation. 
With local $h$-isomorphism, \textit{all our previous conclusions based on standard isomorphism still apply}. For example, GAE (without node-discriminating features) still cannot discriminate locally $h$-non-isomorphic links. And a node-most-expressive GNN with labeling trick can learn the most expressive structural representations of node sets w.r.t. local $h$-isomorphism, i.e., learn the same representation for two node sets if and only if they are locally $h$-isomorphic:

\begin{corollary}\label{labeltrickthmlocal}
Given a node-most-expressive \text{\normalfont GNN} and an injective set aggregation function \text{\normalfont AGG}, then for any $S,\tA, S',\tA',h$, $\text{\normalfont GNN}(S,\tA^{(S)}_{(S,h)}) = \text{\normalfont GNN}(S',\tA'^{(S')}_{(S',h)}) \Leftrightarrow (S,\tA_{(S,h)}) \!\simeq\! (S',\tA'_{(S',h)})$.
\end{corollary}

Corollary~\ref{labeltrickthmlocal} demonstrates labeling trick's power in the context of local isomorphism. To switch to local $h$-isomorphism, all we need to do is to extract the $h$-hop enclosing subgraph around a node set, and apply labeling trick and GNN only to the extracted subgraph. This is exactly what SEAL does. 

\section{Related work}
There is emerging interest in studying the representation power of graph neural networks recently. \citet{xu2018powerful} and \citet{morris2019weisfeiler} first show that the discriminating power of GNNs performing neighbor aggregation is bounded by the 1-WL test. Many works have since been proposed to increase the power of GNNs by simulating higher-order WL tests~\citep{morris2019weisfeiler,maron2019provably,chen2019equivalence}. However, most previous works focus on improving GNN's whole-graph representation power. Little work has been done to analyze GNN's node/link representation power. 
\citet{Srinivasan2020On} first formally studied the difference between structural representations of nodes and links. Although showing that structural node representations of GNNs cannot perform link prediction, their way to learn structural link representations is to give up GNNs and instead use Monte Carlo samples of node embeddings learned by network embedding methods. In this paper, we show that GNNs combined with labeling trick can as well learn structural link representations, which reassures using GNNs for link prediction. 

Many works have implicitly assumed that if a model can learn node representations well, then combining the pairwise node representations can also lead to good link representations~\citep{grover2016node2vec,kipf2016variational,hamilton2017inductive}. However, we argue in this paper that simply aggregating node representations fails to discriminate a large number of non-isomorphic links, and with labeling trick the aggregation of structural node representations leads to structural link representations. \citet{li2020distance} proposed distance encoding (DE), whose implementations based on $S$-discriminating distances can be shown to be specific labeling tricks. They proved that DE can improve 1-WL-GNNs' discriminating power, enabling them to differentiate almost all $(S,\tA)$ tuples sampled from $r$-regular graphs. Our paper contributes to an important aspect that \citet{li2020distance} overlooked: 1) Our theory focuses on the gap between a GNN's single-node and multi-node representation power. We show even a GNN has maximum node representation power, it still fails to learn structural representations of node sets unless combined with a labeling trick. However, the theory of DE cannot explain this. 2) Our theory is not restricted to $r$-regular graphs, but applies to any graphs. 3) Our theory points out that a valid labeling trick is not necessarily distance based---it need only be permutation equivariant and $S$-discriminating. More discussion on the difference between DE's theory and the theory in this paper is given in Appendix~\ref{DEdiscussion}.

\citet{you2019position} also noticed that structural node representations of GNNs cannot capture the dependence (in particular distance) between nodes. To learn position-aware node embeddings, they propose P-GNN, which randomly chooses some anchor nodes and aggregates messages only from the anchor nodes. In P-GNN, nodes with similar distances to the anchor nodes, instead of nodes with similar neighborhoods, have similar embeddings. Thus, P-GNN cannot learn structural node/link representations. 
P-GNN also cannot scale to large datasets. \citet{you2021identity} later proposed ID-GNN. As discussed in Section~\ref{sec:discussion}, ID-GNN's node coloring scheme can be seen as a partial labeling trick.

Finally, although labeling trick is formally defined in this paper, various forms of specific labeling tricks have already been used in previous works. To our best knowledge, SEAL~\citep{zhang2018link} proposes the first labeling trick, which is designed to improve GNN's link prediction power. It is later adopted in inductive knowledge graph completion~\citep{teru2020inductive} and matrix completion~\citep{Zhang2020Inductive}, and is generalized into DE~\citep{li2020distance} which works for $|S|>2$ cases. \citet{wan2021principled} use labeling trick for hyperedge prediction.






\section{Experiments}

In this section, we use a two-node task, link prediction, to empirically validate the effectiveness of labeling trick for multi-node representation learning. We use four link existence prediction datasets in Open Graph Benchmark (OGB)~\citep{hu2020open}: \texttt{ogbl-ppa}, \texttt{ogbl-collab},  \texttt{ogbl-ddi}, and \texttt{ogbl-citation2}. These datasets are open-sourced, large-scale (up to 2.9M nodes and 30.6M edges), adopt realistic train/validation/test splits, and have standard evaluation procedures, thus providing an ideal place to benchmark an algorithm's realistic link prediction power. The evaluation metrics include Hits@$K$ and MRR. Hits@$K$ counts the ratio of positive edges ranked at the K-th place or above against all the negative edges. MRR (Mean Reciprocal Rank) computes the reciprocal rank of the true target node against 1,000 negative candidates, averaged over all the true source nodes. Both metrics are higher the better. 
We include more details and statistics of these datasets in Appendix~\ref{detaileddatasets}. Our code is available at \url{https://github.com/facebookresearch/SEAL_OGB}.


\textbf{Baselines.} We use the following baselines for comparison. We use 5 non-GNN methods: CN (common neighbor), AA (Adamic-Adar), MLP, MF (matrix factorization) and Node2vec. Among them, CN and AA are two simple link prediction heuristics based on counting common neighbors, which are used for sanity checking. We use 3 plain GAE baselines: GraphSAGE~\citep{hamilton2017inductive}, GCN~\citep{kipf2016variational}, and GCN+LRGA~\citep{puny2020graph}. These methods use the Hadamard product of pairwise node representations output by a GNN as link representations, without using a labeling trick. Finally, we compare 3 GNN methods using labeling tricks: GCN+DE~\citep{li2020distance}, GCN+DRNL, and SEAL~\citep{zhang2018link}. GCN+DE/GCN+DRNL enhance GCN with the DE/DRNL labeling trick. SEAL uses a GCN and the DRNL labeling trick, with an additional subgraph-level readout SortPooling~\citep{zhang2018end}. More details are in Appendix~\ref{detailedbaselines}. Moreover, we test the zero-one labeling trick in our ablation experiments. Results can be found in Appendix~\ref{ablation}.

\textbf{Results and discussion.} We present the main results in Table~\ref{ogb3}. Firstly, we can see that GAE methods without labeling trick do not always outperform non-GNN methods. For example, on \texttt{ogbl-ppa} and \texttt{ogbl-collab}, simple heuristics CN and AA outperform plain GAE methods by large margins. This suggests that GAE methods cannot even learn simple neighborhood-overlap-based heuristics, verifying our argument in Introduction. In contrast, when GNNs are enhanced by labeling trick, they are able to beat heuristics. With labeling trick, GNN methods achieve new state-of-the-art performance on 3 out of 4 datasets. In particular, we observe that SEAL outperforms GAE and positional embedding methods, sometimes by surprisingly large margins. For example, in the challenging \texttt{ogbl-ppa} graph, SEAL achieves an Hits@100 of 48.80, which is \textbf{87\%-195\% higher} than GAE methods without using labeling trick. On \texttt{ogbl-ppa}, \texttt{ogbl-collab} and \texttt{ogbl-citation2}, labeling trick methods also achieve state-of-the-art results.



\begin{table*}[t]
\caption{Results for \texttt{ogbl-ppa}, \texttt{ogbl-collab}, \texttt{ogbl-ddi} and \texttt{ogbl-citation2}.}
\vspace{-10pt}
\label{ogb3}
\begin{center}
  \resizebox{0.99\textwidth}{!}{
  \begin{tabular}{llcccccccc}
    \toprule
    &&\multicolumn{2}{c}{\texttt{ogbl-ppa}} & \multicolumn{2}{c}{\texttt{ogbl-collab}} &\multicolumn{2}{c}{\texttt{ogbl-ddi}} & \multicolumn{2}{c}{\texttt{ogbl-citation2}} \\
    &&\multicolumn{2}{c}{Hits@100 (\%)} & \multicolumn{2}{c}{Hits@50 (\%)} & \multicolumn{2}{c}{Hits@20 (\%)} & \multicolumn{2}{c}{MRR (\%)}\\
    \cmidrule(r{0.5em}){3-4} \cmidrule(l{0.5em}){5-6} \cmidrule(l{0.5em}){7-8} \cmidrule(l{0.5em}){9-10}
    \textbf{Category}&\textbf{Method}&Validation&\textbf{Test}&Validation&\textbf{Test}&Validation&\textbf{Test}&Validation&\textbf{Test}  \\
    \midrule
    \multirow{5}{*}{Non-GNN} 
    &\textbf{CN} & 28.23{\small$\pm$0.00} &  27.6{\small$\pm$0.00} & 60.36{\small$\pm$0.00} & 61.37{\small$\pm$0.00} & 9.47{\small$\pm$0.00} & 17.73{\small$\pm$0.00} & 51.19{\small$\pm$0.00} & 51.47{\small$\pm$0.00}\\
    &\textbf{AA} & 32.68{\small$\pm$0.00} &  32.45{\small$\pm$0.00} & 63.49{\small$\pm$0.00} & 64.17{\small$\pm$0.00} & 9.66{\small$\pm$0.00} & 18.61{\small$\pm$0.00} & 51.67{\small$\pm$0.00} & 51.89{\small$\pm$0.00}\\
    &\textbf{MLP} & 0.46{\small$\pm$0.00} &  0.46{\small$\pm$0.00} & 24.02{\small$\pm$1.45} & 19.27{\small$\pm$1.29} & -- & -- & 29.03{\small$\pm$0.17} & 29.06{\small$\pm$0.16}\\
    &\textbf{Node2vec} & 22.53{\small$\pm$0.88} &  22.26{\small$\pm$0.88} & 57.03{\small$\pm$0.52} & 48.88{\small$\pm$0.54} & 32.92{\small$\pm$1.21} &  23.26{\small$\pm$2.09} & 61.24{\small$\pm$0.11} & 61.41{\small$\pm$0.11} \\
    &\textbf{MF} & 32.28{\small$\pm$4.28} &  32.29{\small$\pm$0.94} & 48.96{\small$\pm$0.29} & 38.86{\small$\pm$0.29} & 33.70{\small$\pm$2.64} &  13.68{\small$\pm$4.75} & 51.81{\small$\pm$4.36} & 51.86{\small$\pm$4.43} \\
    \midrule
    \multirow{3}{*}{Plain GAE} 
    &\textbf{GraphSAGE} & 17.24{\small$\pm$2.64} &  16.55{\small$\pm$2.40} & 56.88{\small$\pm$0.77} & 54.63{\small$\pm$1.12} & \cellcolor[HTML]{D2FFFC}62.62{\small$\pm$0.37} &  \cellcolor[HTML]{D2FFFC}53.90{\small$\pm$4.74} & \cellcolor[HTML]{E7FFFD}82.63{\small$\pm$0.23} & \cellcolor[HTML]{E7FFFD}82.60{\small$\pm$0.36} \\
    &\textbf{GCN} & 18.45{\small$\pm$1.40} &  18.67{\small$\pm$1.32} & 52.63{\small$\pm$1.15} & 47.14{\small$\pm$1.45} & \cellcolor[HTML]{E7FFFD}55.50{\small$\pm$2.08} &  \cellcolor[HTML]{E7FFFD}37.07{\small$\pm$5.07} & \cellcolor[HTML]{D2FFFC}84.79{\small$\pm$0.23} & \cellcolor[HTML]{D2FFFC}84.74{\small$\pm$0.21} \\
    &\textbf{GCN+LRGA} & 25.75{\small$\pm$2.82} &  26.12{\small$\pm$2.35} & 60.88{\small$\pm$0.59} & 52.21{\small$\pm$0.72} & \cellcolor[HTML]{B7FFFA}\textbf{66.75}{\small$\pm$0.58} & \cellcolor[HTML]{B7FFFA}\textbf{62.30}{\small$\pm$9.12} & 66.48{\small$\pm$1.61} & 66.49{\small$\pm$1.59} \\
    \midrule
    \multirow{3}{*}{Labeling Trick} 
    &\textbf{GCN+DE} & \cellcolor[HTML]{E7FFFD}36.31{\small$\pm$3.59} &  \cellcolor[HTML]{E7FFFD}36.48{\small$\pm$3.78} & \cellcolor[HTML]{E7FFFD}64.13{\small$\pm$0.16} & \cellcolor[HTML]{D2FFFC}64.44{\small$\pm$0.29} & 29.85{\small$\pm$2.25} &  26.63{\small$\pm$6.82} & 60.17{\small$\pm$0.63} & 60.30{\small$\pm$0.61} \\
    &\textbf{GCN+DRNL} & \cellcolor[HTML]{D2FFFC}46.43{\small$\pm$3.03} &  \cellcolor[HTML]{D2FFFC}45.24{\small$\pm$3.95} & \cellcolor[HTML]{D2FFFC}64.51{\small$\pm$0.42} & \cellcolor[HTML]{E7FFFD}64.40{\small$\pm$0.45} & 29.47{\small$\pm$1.54} &  22.81{\small$\pm$4.93} & 81.07{\small$\pm$0.30} & 81.27{\small$\pm$0.31} \\
    &\textbf{SEAL} & \cellcolor[HTML]{B7FFFA}\textbf{51.25}{\small$\pm$2.52} &  \cellcolor[HTML]{B7FFFA}\textbf{48.80}{\small$\pm$3.16} & \cellcolor[HTML]{B7FFFA}\textbf{64.95}{\small$\pm$0.43} & \cellcolor[HTML]{B7FFFA}\textbf{64.74}{\small$\pm$0.43} & 28.49{\small$\pm$2.69} &  30.56{\small$\pm$3.86} & \cellcolor[HTML]{B7FFFA}\textbf{87.57}{\small$\pm$0.31} & \cellcolor[HTML]{B7FFFA}\textbf{87.67}{\small$\pm$0.32} \\
  \bottomrule
\end{tabular}
}
\end{center}
\end{table*}

Despite obtaining the best results on three datasets, we observe that labeling trick methods do not perform well on \texttt{ogbl-ddi}. \texttt{ogbl-ddi} is considerably denser than the other graphs. It has 4,267 nodes and 1,334,889 edges, resulting in an average node degree of 500.5. In \texttt{ogbl-ddi}, labeling trick methods fall behind GAE methods using trainable node embeddings. One possible explanation is that \texttt{ogbl-ddi} is so dense that a practical GNN with \textit{limited expressive power} is hard to inductively learn any meaningful structural patterns. In comparison, the transductive way of learning free-parameter node embeddings makes GAEs no longer focus on learning inductive structural patterns, but focus on learning node embeddings. The added parameters also greatly increase GAEs' model capacity.
An interesting future topic is to study how to improve labeling tricks' performance on dense graphs.
Appendix~\ref{ablation} presents more ablation experiments to study the power of different labeling tricks, the effect of subgraph pooling, and the number of hops/layers.

\section{Conclusions}
In this paper, we proposed a theory of using GNNs for multi-node representation learning. We first pointed out the key limitation of a common practice in previous works that directly aggregates node representations as a node set representation. To address the problem, we proposed labeling trick which gives target nodes distinct labels in a permutation equivariant way. We proved that labeling trick enables GNNs to learn most expressive structural representations of node sets, which formally characterizes GNNs' maximum multi-node representation learning ability. Experiments on four OGB datasets verified labeling trick's effectiveness.




\section*{Acknowledgements}
The authors greatly thank the actionable suggestions from the reviewers. Li is partly supported by the 2021 JP Morgan Faculty Award and the National Science Foundation (NSF) award HDR-2117997.




\bibliography{paper}
\bibliographystyle{unsrtnat}

\newpage
\appendix

\setlength{\abovedisplayskip}{7pt plus2pt minus5pt}
\setlength{\belowdisplayskip}{7pt plus2pt minus5pt}

\section{Proof of Theorem~\ref{labeltrickthm}}

We restate Theorem~\ref{labeltrickthm}: Given a node-most-expressive \text{\normalfont GNN} and an injective set aggregation function \text{\normalfont AGG}, for any $S,\tA, S',\tA'$, $\text{\normalfont GNN}(S,\tA^{(S)}) = \text{\normalfont GNN}(S',\tA'^{(S')}) \Leftrightarrow (S,\tA) \!\simeq\! (S',\tA') \nonumber$, where $\text{\normalfont GNN}(S,\tA^{(S)}) := \text{\normalfont AGG}(\{\text{\normalfont GNN}(i,\tA^{(S)}) | i\in S\})$.

\begin{proof}

We need to show $\text{\normalfont AGG}(\{\text{\normalfont GNN}(i,\tA^{(S)})|i\in S\})= \text{\normalfont AGG}(\{\text{\normalfont GNN}(i,\tA'^{(S')}|i\in S'\})$ iff $(S,\tA) \simeq (S',\tA')$.

To prove the first direction, we notice that with an injective AGG, 
\begin{align}
    &\text{\normalfont AGG}(\{\text{\normalfont GNN}(i,\tA^{(S)}))|i\in S\}) = \text{\normalfont AGG}(\{\text{\normalfont GNN}(i,\tA'^{(S')}))|i\in S'\}) \nonumber\\
\Longrightarrow ~& \exists~v_1\in S, v_2\in S',~\text{such that}~\text{\normalfont GNN}(v_1,\tA^{(S)}) = \text{\normalfont GNN}(v_2,\tA'^{(S')}) \\
\Longrightarrow ~& (v_1,\tA^{(S)}) \simeq (v_2,\tA'^{(S')}) ~~~~\text{(because GNN is node-most-expressive)}\\
\Longrightarrow ~&  \exists~\pi \in \Pi_n,~\text{such that}~ v_1 = \pi(v_2), \tA^{(S)} = \pi(\tA'^{(S')}).\label{eq:3}
\end{align}

Remember $\tA^{(S)}$ is constructed by stacking $\tA$ and $\tL^{(S)}$ in the third dimension, where $\tL^{(S)}$ is a tensor satisfying: $\forall \pi \in \Pi_n$, \text{\normalfont (1)} $\tL^{(S)} = \pi(\tL^{(S')}) \Rightarrow S = \pi(S')$, and \text{\normalfont (2)} $S = \pi(S'), \tA = \pi(\tA') \Rightarrow \tL^{(S)} = \pi(\tL^{(S')})$. With $\tA^{(S)} = \pi(\tA'^{(S')})$, we have both 
\begin{align}
    \tA = \pi(\tA'),~\tL^{(S)} = \pi(\tL^{(S')}).\nonumber
\end{align}
Because $\tL^{(S)} = \pi(\tL^{(S')}) \Rightarrow S = \pi(S')$, continuing from Equation~(\ref{eq:3}), we have
\begin{align}
    &\text{\normalfont AGG}(\{\text{\normalfont GNN}(i,\tA^{(S)})|i\in S\}) = \text{\normalfont AGG}(\{\text{\normalfont GNN}(i,\tA'^{(S')})|i\in S'\}) \nonumber\\
\Longrightarrow ~& \exists~\pi \in \Pi_n,~\text{such that}~ \tA = \pi(\tA'),~\tL^{(S)} = \pi(\tL^{(S')}) \\ 
\Longrightarrow ~& \exists~\pi \in \Pi_n,~\text{such that}~ \tA = \pi(\tA'),~S = \pi(S')\\ 
\Longrightarrow ~& (S,\tA) \simeq (S',\tA').
\end{align}

Now we prove the second direction. Because $S = \pi(S'), \tA = \pi(\tA') \Rightarrow \tL^{(S)} = \pi(\tL^{(S')})$, we have:
\begin{align}
    &(S,\tA) \simeq (S',\tA') \nonumber\\
\Longrightarrow ~& \exists~\pi \in \Pi_n,~\text{such that}~ S = \pi(S'), \tA = \pi(\tA') \\
\Longrightarrow ~& \exists~\pi \in \Pi_n,~\text{such that}~ S = \pi(S'), \tA = \pi(\tA'), \tL^{(S)} = \pi(\tL^{(S')}) \\
\Longrightarrow ~& \exists~\pi \in \Pi_n,~\text{such that}~ S = \pi(S'), \tA^{(S)} = \pi(\tA'^{(S')}) \\
\Longrightarrow ~& \exists~\pi \in \Pi_n,~\text{such that}~ \forall v_1\in S, v_2 \in S', v_1 = \pi(v_2), ~\text{we have}~ \text{\normalfont GNN}(v_1,\tA^{(S)}) = \text{\normalfont GNN}(v_2,\tA'^{(S')}) \\
\Longrightarrow ~&  \text{\normalfont AGG}(\{\text{\normalfont GNN}(v_1,\tA^{(S)})|v_1\in S\}) = \text{\normalfont AGG}(\{\text{\normalfont GNN}(v_2,\tA'^{(S')})|v_2\in S'\}),
\end{align}
which concludes the proof.

\end{proof}

\section{Proof of Theorem~\ref{thm:link-aso-num}}
We restate Theorem~\ref{thm:link-aso-num}: In any non-attributed graph with $n$ nodes, if the degree of each node in the graph is between $1$ and $\mathcal{O}(\log^{\frac{1-\epsilon}{2h}} n)$ for any constant $\epsilon>0$, then there exists $\omega(n^{2\epsilon})$ many pairs of non-isomorphic links $(u,w), (v,w)$ such that an $h$-layer \textnormal{1-WL-GNN} gives $u,v$ the same representation, while with labeling trick the \textnormal{1-WL-GNN} gives $u,v$ different representations.

\begin{proof}
Our proof has two steps. First, we would like to show that there are $\omega(n^{\epsilon})$ nodes that are locally $h$-isomorphic (see Definition~\ref{localhiso}) to each other. Then, we prove that among these nodes, there are at least $\omega(n^{2\epsilon})$ pairs of nodes such that there exists another node constructing locally $h$ non-isomorphic links with either of the two nodes in each node pair. 

\textbf{Step 1.} Consider an arbitrary node $v$ and denote the subgraph induced by the nodes that are at most $h$-hop away from $v$ as $G_v^{(h)}$ (the $h$-hop enclosing subgraph of $v$). As each node is with degree $d = \mathcal{O}(\log^{\frac{1-\epsilon}{2h}} n)$, then the number of nodes in $G_v^{(h)}$, denoted by $|V(G_v^{(h)})|$, satisfies 
\begin{align*}
 |V(G_v^{(h)})| \leq \sum_{i=0}^{h} d^i = \mathcal{O}(d^h) = \mathcal{O}(\log^{\frac{1-\epsilon}{2}} n).
\end{align*}
We set the max $K=\max_{v\in V} |V(G_v^{(h)})|$ and thus $K =  \mathcal{O}(\log^{\frac{1-\epsilon}{2}} n)$. 

Now we expand subgraphs $G_v^{(h)}$ to $\bar{G}_v^{(h)}$ by adding $K-|V(G_v^{(h)})|$ independent nodes for each node $v\in V$. Then, all $\bar{G}_v^{(h)}$ have the same number of nodes, which is $K$, though they may not be connected graphs. 

Next, we consider the number of non-isomorphic graphs over $K$ nodes. Actually, the number of non-isomorphic graph structures over $K$ nodes is bounded by $2^{K \choose 2} = \exp(\mathcal{O}(\log^{1-\epsilon} n))= o(n^{1-\epsilon})$.

Therefore, due to the pigeonhole principle, there exist $n/o(n^{1-\epsilon}) = \omega(n^\epsilon)$ many nodes $v$ whose $\bar{G}_v^{(h)}$ are isomorphic to each other. Denote the set of these nodes as $V_{iso}$, which consist of nodes that are all locally $h$-isomorphic to each other. Next, we focus on looking for other nodes to form locally $h$-non-isomorphic links with nodes $V_{iso}$. 

\textbf{Step 2.} Let us partition $V_{iso}=\cup_{i=1}^q V_i$ so that for all nodes in $V_i$, they share the same first-hop neighbor sets. Then, consider any pair of nodes $u,v$ such that $u,v$ are from different $V_i$'s. Since $u,v$ share identical $h$-hop neighborhood structures, an $h$-layer 1-WL-GNN will give them the same representation. Then, we may pick one $u$'s first-hop neighbor $w$ that is not $v$'s first-hop neighbor. We know such $w$ exists because of the definition of $V_i$. As $w$ is $u$'s first-hop neighbor and is not $v$'s first-hop neighbor, $(u,w)$ and $(v,w)$ are not isomorphic. With labeling trick, the $h$-layer 1-WL-GNN will give $u,v$ different representations immediately after the first message passing round due to $w$'s distinct label. Therefore, we know such a $(u,w), (v,w)$ pair is exactly what we want.

Based on the partition $V_{iso}$, we know the number of such non-isomorphic link pairs $(u, w)$ and $(v,w)$ is at least:
\begin{align}\label{eq:apd1}
    Y \geq \prod_{i,j=1, i\neq j}^q |V_i||V_j| = \frac{1}{2}\left[(\sum_{i=1}^q|V_i|)^2 - \sum_{i=1}^q|V_i|^2\right].
\end{align}

Because of the definitions of the partition,  $\sum_{i=1}^q|V_i| = |V_{iso}|=\omega(n^\epsilon)$ and the size of each $V_i$ satisfies
\begin{align*}
    1\leq  |V_i| \leq d_w = \mathcal{O}(\log^{\frac{1-\epsilon}{2h}} n),
\end{align*}
where $w$ is one of the common first-hop neighbors shared by all nodes in $V_i$ and $d_w$ is its degree.

By plugging in the range of $|V_i|$, Eq.\ref{eq:apd1} leads to 
\begin{align*}
    Y \geq \frac{1}{2}(\omega(n^{2\epsilon}) - \omega(n^\epsilon)\mathcal{O}(\log^{\frac{1-\epsilon}{2h}} n)) =\omega(n^{2\epsilon}),
\end{align*}
which concludes the proof. 

\end{proof}

\section{Proof of Lemma~\ref{wlconverge}}
We restate Lemma~\ref{wlconverge}: Given a graph with $n$ nodes, a 1-WL-GNN takes up to $\mathcal{O}(n)$ message passing layers to discriminate all the nodes that 1-WL can discriminate.

\begin{proof}
We first note that after one message passing layer, 1-WL-GNN gives different embeddings to any two nodes that 1-WL gives different colors to after one iteration. So we only need to show how many iterations 1-WL takes to converge in any graph.

Note that if two nodes are given different colors by 1-WL at some iteration (they are discriminated by 1-WL), their colors are always different in any future iteration. And if at some iteration, all nodes' colors are the same as their colors in the last iteration, then 1-WL will stop (1-WL fails to discriminate any more nodes and has converged). Therefore, before termination, 1-WL will increase its total number of colors by at least 1 after every iteration. Because there are at most $n$ different final colors given an $n$-node graph, 1-WL takes at most $n-1=\mathcal{O}(n)$ iterations before assigning all nodes different colors.

Now it suffices to show that there exists an $n$-node graph that 1-WL takes $\mathcal{O}(n)$ iterations to converge. Suppose there is a path of $n$ nodes. Then by simple calculation, it takes $\lceil n/2 \rceil$ iterations for 1-WL to converge, which concludes the proof.
\end{proof}

\section{Comparisons between DRNL and DE}\label{drnlde}

In this section, we discuss the relationships and differences between DRNL~\citep{zhang2018link} and DE~\citep{li2020distance} (using shortest path distance). Although they are theoretically equivalent in the context of link prediction, there are some subtle differences that might result in significant performance differences.

Suppose $x$ and $y$ are the two end nodes of the link. \textbf{DRNL} (Double Radius Node Labeling) always assigns label 1 to $x$ and $y$. Then, for any node $i$ with $(d(i,x), d(i,y)) = (1,1)$, it assigns a label $2$. Nodes with radius $(1,2)$ or $(2,1)$ get label 3. Nodes with radius $(1,3)$ or $(3,1)$ get 4. Nodes with $(2,2)$ get 5. Nodes with $(1,4)$ or $(4,1)$ get 6. Nodes with $(2,3)$ or $(3,2)$ get 7. So on and so forth. In other words, DRNL iteratively assigns larger labels to nodes with a larger radius w.r.t. both the two end nodes. The DRNL label $f_l(i)$ of a node $i$ can be calculated by the following hashing function:
\begin{align}
f_l(i) = 1 + \text{min}(d_x, d_y) + (d / 2)[(d / 2) + (d \% 2) - 1],
\label{hashing}
\end{align}
where $d_x := d(i,x)$, $d_y := d(i,y)$, $d := d_x + d_y$, $(d / 2)$ and $(d \% 2)$ are the integer quotient and remainder of $d$ divided by $2$, respectively. This hashing function allows fast closed-form computations of DRNL labels. For nodes with $d(i,x)=\infty$ or $d(i,y)=\infty$, DRNL assigns them a null label 0. Later, the one-hot encoding of these labels are fed to a GNN as the initial node features, or equivalently, we can feed the raw integer labels to an embedding layer first.

Instead of encoding $(d(i,x), d(i,y))$ into a single integer label, \textbf{DE} (distance encoding) directly uses the vector $[d(i,x), d(i,y)]$ as a size-2 label for node $i$. Then, these size-2 labels will be transformed to two-hot encoding vectors to be used as the input node features to GNN. Equivalently, we can also input the size-2 labels to an embedding layer and use the sum-pooled embedding as the initial node features.

These two ways of encoding $(d(i,x), d(i,y))$ theoretically have the same expressive power. However, DRNL and DE have some subtle differences in their implementations. The \textbf{first difference} is that DE sets a maximum distance $d_{\text{max}}$ (a small integer such as 3) for each $d(i,x)$ or $d(i,y)$, i.e., if $d(i,x) \geq d_{\text{max}}$, DE will let $d(i,x) = d_{\text{max}}$. This potentially can avoid some overfitting by reducing the number of possible DE labels as claimed in the original paper~\citep{li2020distance}.

The \textbf{second difference} is that when computing the distance $d(i,x)$, DRNL will temporarily mask node $y$ and all its edges, and when computing the distance $d(i,y)$, DRNL will temporarily mask node $x$ and all its edges.
The reason for this ``masking trick'' is because DRNL aims to use the pure distance between $i$ and $x$ without the influence of $y$. If we do not mask $y$, $d(i,x)$ will be upper bounded by $d(i,y) + d(x,y)$, which obscures the ``true distance'' between $i$ and $x$ and might hurt the node labels' ability to discriminate structurally-different nodes. As we will show in Appendix~\ref{ablation}, this masking trick has a great influence on the performance, which explains DE's inferior performance than DRNL in our experiments.

As we will show in Table~\ref{ogb3}, DRNL has significantly better performance than DE on some datasets. To study what is the root cause for these in-principle equivalent methods's different practical performance, we propose \textbf{DE$^+$}, which adopts DRNL's masking trick in DE. We also try to not set a maximum distance in DE$^+$. This way, there are no more differences in terms of the expressive power between DE$^+$ and DRNL. And we indeed observed that DE$^+$ is able to catch up with DRNL in those datasets where DE does not perform well, as we will show in Appendix~\ref{ablation:de}.

\section{More discussion on the differences between DE's theory and ours}\label{DEdiscussion}
Inspired by the empirical success of SEAL~\citep{zhang2018link}, \citet{li2020distance} proposed distance encoding (DE). It generalizes SEAL's distance-based node labeling (DRNL) for link prediction to arbitrary node set prediction, and theoretically studies how the distance information improves 1-WL-GNN's discriminating power. The main theorem in \citep{li2020distance} (Theorem 3.3) proves that under mild conditions, a 1-WL-GNN combined with DE can discriminate any $(S,\mA),(S',\mA')$ pair sampled uniformly from all $r$-regular graphs, with high probability. This is a significant result, as 1-WL-GNN's discriminating power is bounded by 1-WL, which fails to discriminate any nodes or node sets from $r$-regular graphs. DE's theory shows that with DE we can break the limit of 1-WL and 1-WL-GNN on this major class of graphs where without DE they always fail. 

Despite the success, DE's theory also has several limitations. Firstly, its analysis focuses on the space of random graphs (in particular regular graphs that 1-WL-GNNs fail to represent well). Secondly, DE's theory does not answer whether a GNN combined with DE can learn structural representations, which are the core for joint node set prediction tasks such as link prediction according to \citep{Srinivasan2020On}. Thirdly, although DE's definition (Definition 3.1 of \citep{li2020distance}) only requires permutation invariance, its theory and practical implementations require distance-based node labeling. It is unknown whether other node labeling tricks (including those do not rely on distance) are also useful.


Our theory partly addresses these limitations and is orthogonal to DE's theory, as: 1) We define labeling trick, which is not necessarily distance-based. We show a valid labeling trick need only be permutation equivariant and target-node-set-discriminating. 2) We show with a sufficiently expressive GNN, labeling trick enables learning structural representations of node sets, answering the open question in \citep{Srinivasan2020On} which DE's theory fails to address. 3) We show labeling trick's usefulness for arbitrary graphs, instead of only regular graphs.

Nevertheless, we are uncertain whether DE's power for regular graphs can transfer to any valid labeling trick, including those not based on distance. Thus, we leave an open question here for future research: whether an arbitrary labeling trick enables a 1-WL-GNN to discriminate any $(S,\mA),(S',\mA')$ pair sampled uniformly from all $r$-regular graphs, with high probability? Our guess is that the answer is yes for $|S|=1$ and $|S|=2$. This is because, with an injective message passing layer, we can propagate the unique labels of $S$ to other nodes, thus ``recovering'' the distance information through iterative message passing. We leave a rigorous proof or disproof to future work.

\section{More details about the datasets}\label{detaileddatasets}

We compare the link prediction performance of different baselines on \texttt{ogbl-ppa}, \texttt{ogbl-collab},  \texttt{ogbl-ddi}, and \texttt{ogbl-citation2}. Among them, \texttt{ogbl-ppa} is a protein-protein association graph where the task is to predict biologically meaningful associations between proteins. \texttt{ogbl-collab} is an author collaboration graph, where the task is to predict future collaborations. \texttt{ogbl-ddi} is a drug-drug interaction network, where each edge represents an interaction between drugs which indicates the joint effect of taking the two drugs together is considerably different from their independent effects. \texttt{ogbl-citation2} is a paper citation network, where the task is to predict missing citations. We present the statistics of these OGB datasets in Table~\ref{ogb}. More information about these datasets can be found in \citep{hu2020open}.

\begin{table*}[h]
\caption{Statistics and evaluation metrics of OGB link prediction datasets.}
\label{ogb}
\begin{center}
  \resizebox{0.85\textwidth}{!}{
  \begin{tabular}{lcccccc}
    \toprule
    \textbf{Dataset}&\textbf{\#Nodes}&\textbf{\#Edges}&\textbf{Avg. node deg.}&\textbf{Density}&\textbf{Split ratio}&\textbf{Metric} \\
    \midrule
    \texttt{ogbl-ppa} & 576,289 &  30,326,273 & 73.7 & 0.018\% & 70/20/10 & Hits@100\\
    \texttt{ogbl-collab} & 235,868 & 1,285,465 & 8.2 & 0.0046\% & 92/4/4 & Hits@50\\
    \texttt{ogbl-ddi} & 4,267 & 1,334,889 & 500.5 & 14.67\% & 80/10/10 & Hits@20\\
    \texttt{ogbl-citation2} & 2,927,963 & 30,561,187 & 20.7 & 0.00036\% & 98/1/1 & MRR\\
  \bottomrule
\end{tabular}
}
\end{center}
\end{table*}

We choose OGB datasets for benchmarking our methods because these datasets adopt realistic train/validation/test splitting methods, such as by resource cost in laboratory (\texttt{ogbl-ppa}), by time (\texttt{ogbl-collab} and \texttt{ogbl-citation2}), and by drug target in the body (\texttt{ogbl-ddi}). They are also large-scale (up to 2.9M nodes and 30.6M edges), open-sourced, and have standard evaluation metrics. OGB has an official leaderboard\footnote{\url{https://ogb.stanford.edu/docs/leader\_linkprop/}}, too, providing a place to fairly compare different methods' link prediction performance.

\section{More details about the baselines} \label{detailedbaselines}

We include baselines achieving top places on the OGB leaderboard. All the baselines have their open-sourced code and paper available from the leaderboard. We adopt the numbers published on the leaderboard if available, otherwise we run the method ourselves using the open-sourced code. Note that there are other potentially strong baselines that we have to omit here, because they cannot easily scale to OGB datasets. For example, we have contacted the authors of P-GNN~\citep{you2019position}, and confirmed that P-GNN is not likely to scale to OGB datasets due to the computation of all-pairs shortest path distances.

All the compared methods are in the following. We briefly describe how each method obtains its final node representations.

\begin{itemize}[leftmargin=15pt,itemsep=-1pt,topsep=-2pt]
\item \textbf{MLP}: Node features are directly used as the node representations without considering graph structure.
\item \textbf{Node2vec}~\citep{perozzi2014deepwalk,grover2016node2vec}: The node representations are the concatenation of node features and Node2vec embeddings.
\item \textbf{MF} (Matrix Factorization): Use free-parameter node embeddings trained end-to-end as the node representations.
\item \textbf{GraphSAGE}~\citep{hamilton2017inductive}: A GAE method with GraphSAGE as the GNN.
\item \textbf{GCN}~\citep{kipf2016variational}: A GAE method with GCN as the GNN.
\item \textbf{LRGA}~\citep{puny2020graph}: A GAE method with LRGA-module-enhanced GCN.
\item \textbf{GCN+DE}: Apply GCN to the DE~\citep{li2020distance} labeled graphs.
\item \textbf{GCN+DRNL}: Apply GCN to the DRNL~\citep{zhang2018link} labeled graphs.
\item \textbf{SEAL}~\citep{zhang2018link}: The same as GCN+DRNL with an additional subgraph-level readout. Note that we reimplemented SEAL in this paper with a greatly improved efficiency and flexibility than the original implementation\footnote{\url{https://github.com/muhanzhang/SEAL}}. The code will be released in the future.
\end{itemize}

Except SEAL, all models use the Hadamard product between pairwise node representations as the link representations. The link representations are fed to an MLP for final prediction. All the GAE methods' GNNs have 3 message passing layers with 256 hidden dimensions, with a tuned dropout ratio in $\{0,0.5\}$. All the labeling trick methods (GCN+DE, GCN+DRNL and SEAL) extract 1-hop enclosing subgraphs. The GCNs in GCN+DRNL and GCN+DE also use 3 message passing layers with 256 hidden dimensions for consistency. The GNN in SEAL follows the DGCNN in the original paper, which has 3 GCN layers with 32 hidden dimensions each, plus a SortPooling layer~\citep{zhang2018end} and several 1D convolution layers after the GCN layers to readout the subgraph. The use of a subgraph-level readout instead of only reading out two nodes is not an issue for SEAL, because 1) the two center nodes' information is still included in the output of the subgraph-level readout, and 2) the inclusion of additional neighborhood node representations may help learn better neighborhood features than only reading out two center nodes. As we will show in Appendix~\ref{subgaph_readout}, a subgraph-level readout sometimes improves the performance.

The \texttt{ogbl-ddi} graph contains no node features, so MLP is omitted, and the GAE methods here use free-parameter node embeddings as the GNN input node features and train them together with the GNN parameters. For labeling trick methods, the node labels are input to an embedding layer and then concatenated with the node features (if any) as the GNN input. Note that the original SEAL can also include pretrained node embeddings as additional features. But according to \citep{Srinivasan2020On}, node embeddings bring no additional value given structural representations. This is also consistent with our observation and the experimental results of~\citep{zhang2018link}, where including node embeddings gives no better results. Thus, we give up node embeddings in SEAL.


For the baseline GCN+LRGA, its default hyperparameters result in out of GPU memory on \texttt{ogbl-citation2}, even we use an NVIDIA V100 GPU with 32GB memory. Thus, we have to reduce its hidden dimension to 16 and matrix rank to 10. It is possible that it can achieve better performance with a larger hidden dimension and larger matrix rank using a GPU with a larger memory.

We implemented the labeling trick methods (GCN+DE, GCN+DRNL and SEAL) using the PyTorch Geometric~\citep{fey2019fast} package. For all datasets, labeling trick methods only used a fixed 1\% to 10\% of all the available training edges as the positive training links, and sampled an equal number of negative training links randomly. Labeling trick methods showed excellent performance even without using the full training data, which indicates its strong inductive learning ability. Due to using different labeled subgraphs for different links, labeling trick methods generally take longer running time than GAE methods. On the largest \texttt{ogbl-citation2} graph, SEAL takes about 7 hours to finishing its training of 10 epochs, and takes another 28 hours to evaluate the validation and test MRR each. For \texttt{ogbl-ppa}, SEAL takes about 20 hours to train for 20 epochs and takes about 4 hours for evaluation. The other two datasets are finished within hours. 

\section{Ablation study}\label{ablation}
In this section, we conduct several ablation experiments to more thoroughly study the effect of different components around labeling trick on the final link prediction performance.

\subsection{How powerful is the zero-one labeling trick?}
Firstly, we aim to understand how powerful the proposed zero-one labeling (Definition~\ref{zolabeling}) is. Although zero-one labeling is a also valid labeling trick that theoretically enables a node-most-expressive GNN to learn structural representations, in practice our GNNs may not be expressive enough. Then how will the zero-one labeling trick perform compared to those more sophisticated ones such as DE and DRNL? We conduct experiments on \texttt{ogbl-collab} and \texttt{ogbl-citation2} to answer this question. In Table~\ref{table:zo}, we compare GCN (1-hop) + all-zero labeling (not a valid labeling trick), GCN (1-hop) + zero-one labeling trick, and GCN (1-hop) + DRNL. All methods use the same 3 GCN layers with 256 hidden dimensions, 1-hop enclosing subgraphs, and Hadamard product of the two end node representations as the link representations. All the remaining settings follow those of GCN+DRNL.

\begin{table}[h]
\caption{Ablation study on the power of the zero-one labeling trick.}
\label{table:zo}
\begin{center}
  \resizebox{0.7\textwidth}{!}{
  \begin{tabular}{lcccc}
    \toprule
    &\multicolumn{2}{c}{\texttt{ogbl-collab}} & \multicolumn{2}{c}{\texttt{ogbl-citation2}} \\
    &\multicolumn{2}{c}{Hits@50 (\%)} & \multicolumn{2}{c}{MRR (\%)} \\
    \cmidrule(r{0.5em}){2-3} \cmidrule(l{0.5em}){4-5}
    \textbf{Method}&Validation&\textbf{Test}&Validation&\textbf{Test}\\
    \midrule
    \textbf{GCN (1-hop) + all-zero labeling} & {24.35\small$\pm$1.28} &  25.92{\small$\pm$1.47} & 36.97{\small$\pm$0.56} & 36.98{\small$\pm$0.57}\\
    \textbf{GCN (1-hop) + zero-one labeling trick} & 44.45{\small$\pm$1.39} &  44.79{\small$\pm$1.26} & 38.73{\small$\pm$0.86} & 38.78{\small$\pm$0.88} \\
    \textbf{GCN (1-hop) + DRNL} & 64.51{\small$\pm$0.42} &  64.40{\small$\pm$0.45} & 81.07{\small$\pm$0.30} & 81.27{\small$\pm$0.31} \\
    \midrule
    \textbf{GIN (1-hop) + zero-one labeling trick} & 60.31{\small$\pm$0.81} &  59.48{\small$\pm$1.17} & 78.32{\small$\pm$1.07} & 78.50{\small$\pm$1.08} \\
  \bottomrule
\end{tabular}
}
\end{center}
\end{table}

From Table~\ref{table:zo}, we can see that GCN+zero-one labeling trick indeed  has better performance than GCN without labeling trick, which aligns with our theoretical results that even a simple zero-one labeling is also a valid labeling trick that enables learning structural representations. Nevertheless, the zero-one labeling trick is indeed less powerful than DRNL, as shown by the performance gaps especially on the \texttt{ogbl-citation2} dataset. We are then interested in figuring out what could cause such large performance differences between two (both valid) labeling tricks, because as Theorem~\ref{labeltrickthm} shows, any valid labeling trick can enable a node-most-expressive GNN to learn structural link representations.

We suspect that the insufficient expressive power of GCN is the cause. Therefore, we change GCN to Graph Isomorphism Network (GIN)~\citep{xu2018powerful}. By replacing the linear feature transformations in GCN with MLPs, GIN is one of the most expressive GNNs based on message passing. The results are shown in the last column of Table~\ref{table:zo}. As we can see, GIN (1-hop) + zero-one labeling trick has much better performance than GCN (1-hop) + zero-one labeling trick, and is almost catching up with GCN (1-hop) + DRNL. The results very well align with our theory---as long as we have a sufficiently expressive GNN, even a simple zero-one labeling trick can be very powerful in terms of enabling learning structural representations. Nevertheless, in practice when we only have less powerful GNNs, we should better choose those more sophisticated labeling tricks such as DE and DRNL for better link prediction performance.

\subsection{DE vs. \textbf{DE$^+$} vs. DRNL}\label{ablation:de}
In Appendix~\ref{drnlde}, we have discussed the differences of the implementations of DE and DRNL. That is, although DE and DRNL are equivalent in theory, there are two differences in their implementations: 1) DE sets a maximum distance $d_\text{max}$ (by default 3) while DRNL does not, and 2) DRNL masks the other end node when computing the distances to one end node and vice versa, while DE does not. To study whether it is these implementation differences between DE and DRNL that result in the large performance differences in Table~\ref{ogb3}, we propose \textbf{DE$^+$} which no longer sets a maximum distance in DE and additionally does the masking trick like DRNL. We compare DE, \text{DE$^+$}, and DRNL on \texttt{ogbl-ppa} and \texttt{ogbl-citation2} (where DE shows significantly lower performance than DRNL in Table~\ref{ogb3}). All of them use GCN as the GNN with the same hyperparameters. The results are shown in Table~\ref{table:de}.

\begin{table}[h]
\caption{Comparison of DE, \text{DE$^+$} and DRNL.}
\label{table:de}
\begin{center}
  \resizebox{0.62\textwidth}{!}{
  \begin{tabular}{lcccc}
    \toprule
    &\multicolumn{2}{c}{\texttt{ogbl-ppa}} & \multicolumn{2}{c}{\texttt{ogbl-citation2}} \\
    &\multicolumn{2}{c}{Hits@100 (\%)} & \multicolumn{2}{c}{MRR (\%)} \\
    \cmidrule(r{0.5em}){2-3} \cmidrule(l{0.5em}){4-5}
    \textbf{Method}&Validation&\textbf{Test}&Validation&\textbf{Test}\\
    \midrule
    \textbf{GCN+DE ($d_\textnormal{max}=3$)} & {36.31\small$\pm$3.59} &  36.48{\small$\pm$3.78} & 60.17{\small$\pm$0.63} & 60.30{\small$\pm$0.61}\\
    \textbf{CCN+DE$^+$ ($d_\textnormal{max}=3$)} & 47.17{\small$\pm$1.84} &  45.70{\small$\pm$3.46} & 74.75{\small$\pm$1.18} & 75.00{\small$\pm$1.20} \\
    \textbf{CCN+DE$^+$ ($d_\textnormal{max}=\infty$)} & 45.81{\small$\pm$3.53} &  43.88{\small$\pm$5.18} & 79.37{\small$\pm$4.50} & 78.85{\small$\pm$0.17} \\
    \textbf{GCN+DRNL} & 46.43{\small$\pm$3.03} &  45.24{\small$\pm$3.95} & 81.07{\small$\pm$0.30} & 81.27{\small$\pm$0.31} \\
  \bottomrule
\end{tabular}
}
\end{center}
\end{table}

We can observe that DE$^+$ outperforms DE by large margins. This indicates that the masking trick used in DRNL is very important. Intuitively, temporarily masking the target node $y$ when computing distances to the source node $x$ can give more diverse node labels. Without the masking, $d(i,x)$ will be upper bounded by $d(i,y) + d(x,y)$. Because the distance between $x$ and $y$ can be small in positive links, without the masking $d(i,x)$ will be restricted to small numbers, which hurts their ability to detect subtle differences between nodes' relative positions within the subgraph. Nevertheless, the benefit of the masking trick is not observed in smaller datasets such as \texttt{ogbl-collab} (Table~\ref{ogb3}). 

We can also find that DE$^+$ without setting a maximum distance has very close performance to DRNL, which aligns with our discussion in Appendix~\ref{drnlde}. By removing the maximum distance restriction, DE$^+$ essentially becomes DRNL. However, there are still small performance differences, possibly because DRNL has a larger embedding table than DE$^+$ (DRNL's maximum label is larger) which results in a slightly larger model capacity. Nevertheless, this can be alleviated by doubling the embedding dimension of DE$^+$. In summary, we can conclude that the masking trick used in DRNL is crucial to the performance on some datasets. Compared to DE, DE$^+$ and DRNL show better practical performance. Studying more powerful labeling tricks is also an important future direction.

\subsection{Is a subgraph-level readout useful?}\label{subgaph_readout}
In Table~\ref{ogb3}, we observe that SEAL is generally better than GCN+DRNL. SEAL also uses GCN and the DRNL labeling trick, so the main difference is the subgraph-level readout in SEAL. That is, instead of only reading out the two center nodes' representations as the link representation, SEAL performs a readout over all the nodes in the enclosing subgraph. Here we study this effect further by testing whether a subgraph-level sum-pooling readout is also useful. We replace the Hadamard product of two center node representations in GCN+DRNL with the sum over all node representations within the enclosing subgraph. The results are shown in Table~\ref{table:subgraphreadout}.

\begin{table}[h]
\caption{Ablation study on subgraph-level readout.}\label{table:subgraphreadout}
\begin{center}
  \resizebox{0.65\textwidth}{!}{
  \begin{tabular}{lcccc}
    \toprule
    &\multicolumn{2}{c}{\texttt{ogbl-collab}} & \multicolumn{2}{c}{\texttt{ogbl-citation2}} \\
    &\multicolumn{2}{c}{Hits@50 (\%)} & \multicolumn{2}{c}{MRR (\%)} \\
    \cmidrule(r{0.5em}){2-3} \cmidrule(l{0.5em}){4-5}
    \textbf{Method}&Validation&\textbf{Test}&Validation&\textbf{Test}\\
    \midrule
    \textbf{GCN+DRNL} & 64.51{\small$\pm$0.42} &  64.40{\small$\pm$0.45} & 81.07{\small$\pm$0.30} & 81.27{\small$\pm$0.31} \\
    \textbf{GCN+DRNL (sum-pooling)} & 64.64{\small$\pm$0.24} &  63.26{\small$\pm$0.35} & 84.98{\small$\pm$0.23} & 85.20{\small$\pm$0.26} \\
    \textbf{SEAL} & 64.95{\small$\pm$0.43} &  64.74{\small$\pm$0.43} & 87.57{\small$\pm$0.31} & 87.67{\small$\pm$0.32} \\
  \bottomrule
\end{tabular}
}
\end{center}
\end{table}

As we can see, using sum-pooling has a similar effect to the SortPooling in SEAL, i.e., it greatly improves the performance on \texttt{ogbl-citation2}, while slightly reduces the performance on \texttt{ogbl-collab}. This means, using a subgraph-level readout can sometimes be very helpful. Although according to Theorem~\ref{labeltrickthm} we only need to aggregate the representations of the two center nodes (two end nodes of the link) as the link representation, in practice, because our GNNs only have limited expressive power, reading out all nodes within the enclosing subgraph could help GNNs learn better subgraph-level features thus better detecting the target link's local $h$-isomorphism class. Such subgraph representations can be more expressive than only the two center nodes' representations, especially when the number of message passing layers is small so that the center nodes have not gained enough information from the whole subgraph.

\subsection{Is it helpful to make number of layers larger than number of hops?}
In all labeling trick methods, we have used a fixed enclosing subgraph hop number $h=1$, and a fixed number of message passing layers $l=3$. Using a number of message passing layers larger than the number of hops is different from the practice of previous work. For example, in GAE, we always select $h=l$ hops of neighbors if we decide to use $l$ message passing layers. So is it really helpful to use $l>h$? Intuitively, using $l > h$ layers can make GNNs more sufficiently absorb the entire enclosing subgraph information and learn better link representations. Theoretically, as we have shown in Lemma~\ref{wlconverge}, to reach the maximum representation power of 1-WL-GNN, we need to use $\mathcal{O}(n)$ number of message passing layers, where $n$ is the number of nodes in the enclosing subgraph. Thus, using $l>h$ can enhance GNN's representation power and learn more expressive link representations.

\begin{table}[h]
\caption{Ablation study on subgraph-level readout.}\label{table:l>h}
\begin{center}
  \resizebox{0.8\textwidth}{!}{
  \begin{tabular}{lcccccc}
    \toprule
    &\multicolumn{2}{c}{\texttt{ogbl-ppa}}&\multicolumn{2}{c}{\texttt{ogbl-collab}} & \multicolumn{2}{c}{\texttt{ogbl-citation2}} \\
    &\multicolumn{2}{c}{Hits@100 (\%)} &\multicolumn{2}{c}{Hits@50 (\%)} & \multicolumn{2}{c}{MRR (\%)} \\
    \cmidrule(r{0.5em}){2-3} \cmidrule(l{0.5em}){4-5} \cmidrule(l{0.5em}){6-7}
    \textbf{Method}&Validation&\textbf{Test}&Validation&\textbf{Test}&Validation&\textbf{Test}\\
    \midrule
    \textbf{GCN+DRNL $(l=3)$} & 46.43{\small$\pm$3.03} &  45.24{\small$\pm$3.95} & 64.51{\small$\pm$0.42} & 64.40{\small$\pm$0.45} & 81.07{\small$\pm$0.30} & 81.27{\small$\pm$0.31}  \\
    \textbf{GCN+DRNL $(l=1)$} & 31.59{\small$\pm$2.79} &  33.57{\small$\pm$3.06} & 64.38{\small$\pm$0.13} & 63.95{\small$\pm$0.42} & 77.77{\small$\pm$0.42} & 78.02{\small$\pm$0.44} \\
  \bottomrule
\end{tabular}
}
\end{center}
\end{table}

To validate the above , we conduct experiments on GCN+DRNL by using $l=1$ message passing layers (and still $h=1$). The results are shown in Table~\ref{table:l>h}. As we can observe, using $l=1$ results in lower performance than using $l=3$ in all three datasets. On \texttt{ogbl-collab}, this effect is very small. However, on \texttt{ogbl-ppa} and \texttt{ogbl-citation2}, the performance gaps are significant. These results demonstrate the usefulness of using more message passing layers than hops.

Nevertheless, we are unsure whether it is still helpful to make $l>h$ when we use a large $h$, such as $h=2$ or $h=3$. We cannot generally verify this because increasing $h$ will exponentially increase our subgraph sizes. And considering the huge computation cost on two relatively large datasets \texttt{ogbl-ppa} and \texttt{ogbl-citation2}, using $h=1$ is currently the maximum $h$ we can afford. We thus only conduct experiments using different $h$'s on the smallest \texttt{ogbl-collab} dataset. We have tried different combinations of $(l,h)$ from $(1,1)$ all the way up to $(4,3)$, and the testing scores are consistently around 63 to 64. This seems to indicate increasing $h$ or $l$ is not helpful in this dataset. Nevertheless, \texttt{ogbl-collab} may not be representative enough to derive a general conclusion. For example, in the original SEAL paper~\citep{zhang2018link}, the authors found using $h=2$ is helpful for many datasets. Thus, fully answering this question might need further investigations. But when $h=1$, we can conclude that using $l>h$ is better.

\end{document}

%% file: math_commands.tex
\usepackage{amsmath,amsfonts,bm}

\def\eqref#1{equation~\ref{#1}}

\def\1{\bm{1}}

\def\vz{{\bm{z}}}

\def\mA{{\bm{A}}}

\DeclareMathAlphabet{\mathsfit}{\encodingdefault}{\sfdefault}{m}{sl}
\SetMathAlphabet{\mathsfit}{bold}{\encodingdefault}{\sfdefault}{bx}{n}
\newcommand{\tens}[1]{\bm{\mathsfit{#1}}}
\def\tA{{\tens{A}}}

\def\tL{{\tens{L}}}

\def\gG{{\mathcal{G}}}

\newcommand{\R}{\mathbb{R}}

